\documentclass[11pt]{article}

\usepackage[final]{acl}

\usepackage{times}
\usepackage{latexsym}

\usepackage[T1]{fontenc}
\usepackage[utf8]{inputenc}

\usepackage{microtype}

\usepackage{inconsolata}

\usepackage{graphicx}

\usepackage{multirow}
\usepackage{booktabs}
\usepackage{colortbl}
\usepackage{amsmath}
\usepackage{amsfonts,amssymb}
\usepackage{multicol} 
\usepackage{lipsum}
\usepackage{stfloats}
\usepackage{caption}
\usepackage{xcolor} 
\usepackage{pifont}
\newcommand{\tabincell}[2]{\begin{tabular}{@{}#1@{}}#2\end{tabular}}

\definecolor{mygray}{gray}{0.9}
\definecolor{highlight}{RGB}{238,250,215}
\definecolor{lightbluegray}{HTML}{8CBBED}

\usepackage{adjustbox}
\usepackage{tabularx}

\title{Knowing When to Trust Images: Reliability-Aware Multi-modal \\ Entity Alignment}

\author{Chenxiao Li\textsuperscript{1} \;
    Yunhe Feng\textsuperscript{1}\;
    Dongfang Liu\textsuperscript{2}\;
    Dong Nie\textsuperscript{3}\;
    Yan Huang\textsuperscript{1}\;
    Heng Fan\textsuperscript{1} \\
    \textsuperscript{1}University of North Texas
    \;\;
    \textsuperscript{2}Purdue University \;\;
    \textsuperscript{3}Meta Inc\\
    \texttt{chenxiaoli@my.unt.edu} \;\;
    \texttt{heng.fan@unt.edu}
}
\begin{document}
\maketitle
\begin{abstract}

The visual modality, \emph{i.e.}, images, plays a key role in multi-modal entity alignment (MMEA). Existing approaches often directly fuse the image with other modalities to align different entities. Although simple, such strategies overlook the potential noise in the images and their semantic misalignment with corresponding entities, resulting in suboptimal fusion and degraded performance. Addressing this, we propose a novel \emph{\textbf{R}eliability-\textbf{A}ware framework for \textbf{MMEA}} (\textbf{RA-MMEA}), which assesses visual reliability and adaptively improves unreliable visual representations for robust entity alignment. The core lies in two modules, including \emph{dependency-aware visual reliability prediction} (DA-VRP) and \emph{stability-regularized visual embedding generation} (SR-VEG). The former aims to estimate the reliability of an image by leveraging multi-modal dependency within the entity, while the latter focuses on producing alternative visual representation conditioned on semantics encoded in textual modalities for multi-modal fusion. Compared to current methods, RA-MMEA enables more reliable visual representations for modality fusion, thereby improving performance. In extensive experiments, RA-MMEA achieves state-of-the-art results, verifying the importance of reliable visual modality for entity alignment and the effectiveness of RA-MMEA. \emph{The code and results will be released.}

\end{abstract}

\section{Introduction}
\label{Introduction}

\begin{figure}[!t]
    \centering
  \includegraphics[width=1\linewidth]{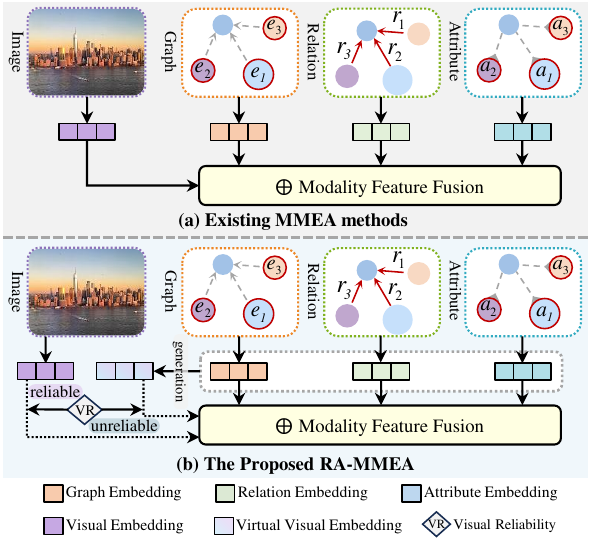}
  \captionsetup{justification=justified, singlelinecheck=false}
  \caption{Comparison of existing methods directly fusing image and other modalities in (a) and RA-MMEA that first assesses the reliability of the image and then performs fusion in (b). \emph{Best viewed in pdf for all figures}.
  }
  \label{fig:question}
\end{figure}

Multi-modal knowledge graphs (MMKGs) employ heterogeneous information such as image, text, and structure to represent real-world entities. Yet, independently constructed MMKGs often contain duplicated or semantically overlapping entities, resulting in fragmented knowledge across graphs. To address this issue, multi-modal entity alignment (MMEA) \citep{chen2020mmea,liu2019mmkg} has been introduced to identify equivalent entities across different MMKGs by leveraging multi-modal information, thus consolidating fragmented MMKGs into unified and complete knowledge resources.

In MMEA, visual modality, \emph{i.e.}, images, offers rich appearance information and is crucial for aligning entities. Existing models typically directly fuse the image and other modalities to construct entity representation (see Fig.~\ref{fig:question} (a)). However, images in MMKGs are \emph{not} always reliable. They may be irrelevant to the entity or exhibit a large semantic gap from the text. Directly incorporating such noisy images into multi-modal fusion may compromise the entity representation and consequently degrade alignment. To alleviate this, a straightforward solution is to filter out unreliable images using manually annotated class conflict dictionaries (CCDs)~\citep{shi2022probing}. However, constructing CCDs entails significant labeling efforts, which is laborious and expensive. Besides, manually designed class conflict rules in CCDs inevitably introduce biases and struggle to generalize to real-world scenarios. % Thus, it is natural to ask: \emph{Can we learn to filter out unreliable images by leveraging information from the entity itself}?

In this paper, we propose to learn visual reliability from the entity itself and adaptively handle unreliable images before multi-modal fusion, and introduce a novel \emph{Reliability-Aware framework for MMEA} (\textbf{RA-MMEA}), that assesses visual
reliability and adaptively improves unreliable
visual representations for robust entity alignment (see Fig.~\ref{fig:question}(b)). The key intuition is that image reliability can be inferred from its consistency with the semantic context of an entity, including graph, relation, and attribute. Based on this intuition, we design a dependency-aware visual reliability predictor (DA-VRP) in RA-MMEA, which estimates image reliability from global multi-modal dependency. Reliable images are expected to form coherent dependency with the remaining entity modalities, whereas unreliable images tend to deviate from such dependency and behave as outliers in the multi-modal representation space. DA-VRP first calculates the dependency score of the image with other modalities through dependency relation modeling, and then uses such dependency score to learn a binary reliability predictor that identifies and suppresses unreliable visual modalities. To enable end-to-end optimization with discrete binary indicators, we employ Gumbel-Softmax to predict the binary reliability in a differentiable manner.

After filtering unreliable images, the suppressed visual information needs to be compensated. To this end, we present a simple yet effective stability-regularized visual embedding generation module in RA-MMEA, termed SR-VEG, that aims to produce alternative visual representation conditioned on other modalities for multi-modal fusion. Specifically, it adaptively integrates features from other modalities and employs latent-variable modeling to generate visual representations. To preserve dependency consistency under modality removal, we further present a weight-difference penalty that aligns dependency scores computed with and without the visual modality. This regularization stabilizes the overall dependency estimation within an entity, leading to better reconstructed visual features and thereby improving entity alignment. 

% To the best of our knowledge, RA-MMEA is the first method that explicitly models visual reliability in MMEA. Compared with existing methods, RA-MMEA enables more reliable visual representations to be exploited for multi-modal fusion, leading to more robust entity alignment. To validate its effectiveness, we conduct extensive experiments on multiple benchmarks. The results show that RA-MMEA achieves state-of-the-art performance, demonstrating the importance of visual reliability in MMEA and the effectiveness of our RA-MMEA.

To the best of our knowledge, RA-MMEA is the first method that explicitly models visual reliability in MMEA. Compared with existing methods, RA-MMEA jointly estimates visual reliability and selectively reconstructs only unreliable visual representations before multi-modal fusion. This reliability-aware paradigm prevents unreliable visual information from degrading entity representations while preserving reliable visual information. Extensive experiments on multiple benchmarks demonstrate that RA-MMEA consistently achieves state-of-the-art performance, highlighting the importance of reliability-aware visual modeling for robust entity alignment.

In summary, our \emph{\textbf{contributions}} are as follows: \ding{171} We propose RA-MMEA, the first reliability-aware framework for MMEA; \ding{170} We present DA-VRP that leverages global multi-modal dependency within an entity for visual reliability prediction; \ding{168} We introduce SR-VEG that generates virtual visual embeddings under stability‑regularized consistency constraints; \ding{169} On multiple benchmarks, our RA-MMEA achieves state-of-the-art results.

\section{Related Work}
\label{sec:Related Work}

\textbf{Multi-modal Entity Alignment (MMEA).} Typical MMEA methods primarily focus on improving multi-modal fusion to optimize embedding representations \citep{zhang2024multimodal}. To mitigate noise arising from modality heterogeneity, existing studies \citep{shi2022probing,li2025exploring} employ annotated ontology matching as a filtering criterion to enhance embedding quality. To handle uncertain missing values in fusion, another line of work \citep{chen2023rethinking,li2023vision,li2024triplet} proposes generating pseudo-representation for the missing modality in multi-modal fusion. In contrast to these strategies that regulate input data quality or quantity, methods such as \citep{lin2022multi,chen2022multi,chen2023meaformer,li2025probing} adopt contrastive learning or Transformer to precisely model inter-modal relationships, thereby alleviating the adverse effects of multi-modal information imbalance. Despite promising results, the aforementioned methods overlook the importance of addressing unreliable visual modality to improve MMEA.

\vspace{0.3em}
\noindent
\textbf{LLMs in Multi-modal Entity Alignment.} Large language models (LLMs) have been recently employed for MMEA due to their strong reasoning and understanding capabilities. \citet{chen2024tackling} employ LLMs to generate and embed attribute abstracts, facilitating attribute alignment. \citet{ji2025capturing} project multi-modal embeddings based on LLM-based attribute summaries to obtain the potential associations between them. \citet{lu2025breaking} propose LLM-guided visual alignment and attribute enhancement to achieve semantic filtering and enhancement in MMEA. \citet{li2026learning} resolve dual-level noisy correspondence via a dedicated two-fold principle and employ MLLMs for relational reasoning. Despite excellent performance, these LLM-based models may introduce potential information leakage by relying on external world knowledge and entity names. Differently, instead of using external knowledge for MMEA, this work focuses on assessing and leveraging the reliability of visual modality based on the entity itself, aiming to improve multi-modal fusion under unreliable images.

\section{Methodology}

\begin{figure*}[htbp]
  \centering
  \includegraphics[width=0.98\textwidth, keepaspectratio]{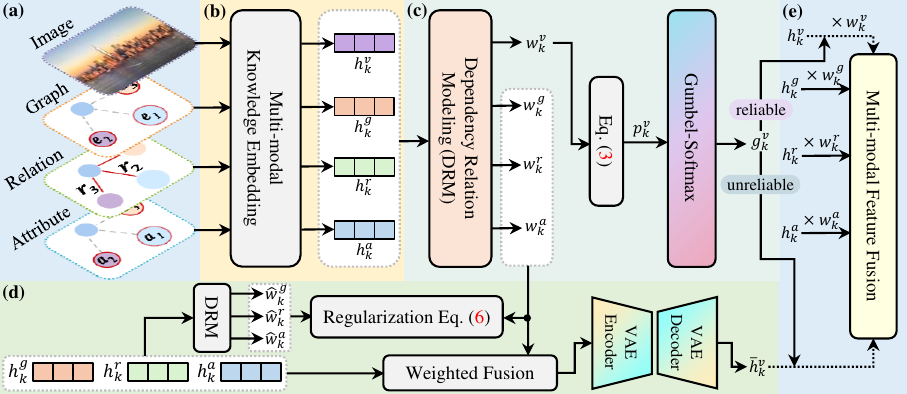}
  \caption{Illustration of the proposed RA-MMEA, which assesses visual modality reliability and adaptively improves unreliable visual representations through generated feature for MMEA. (a) Inputs with image, graph, relation, and attribute; (b) Multi-modal Embedding; (c) Dependency-Aware Visual Reliability Prediction; (d) Stability-Regularized Visual Embedding Generation; and (e) Multi-modal Fusion.
  }
  \label{fig:model}
\end{figure*}

\textbf{Problem Formulation.} In MMEA, a multi-modal knowledge graph (MMKG) is formalized as $G=(E, R, A, T, I, P)$, where $E, R, A, T$, and $I$ are the sets of entities, relations, attributes, triples, and images, and $P=\{(e,i)\mid e\in E,i\in I\}$ denotes the set of entity-image pairs. Each entity is associated to multiple attributes and 0 $\sim$ 1 image. Given two MMKGs $G_{1}=(E_{1}, R_{1}, A_{1}, T_{1}, I_{1}, P_{1})$ and $G_{2}=$ $(E_{2}, R_{2}, A_{2}, T_{2}, I_{2}, P_{2})$, the set of seed alignments across them is defined as $ H=\{(e_{1}, e_{2})\mid e_{1} \in E_{1}, e_{2} \in E_{2}, e_{1} \equiv e_{2}\}$, where $ \equiv $ represents the equivalence of two entities. Given a set of pre-aligned entity pairs as training guidance, MMEA aims to match the counterpart entities $e_{1}$ and $e_{2}$ describing the same concepts in the real world from distinct multi-modal knowledge graphs.

\vspace{0.3em}
\noindent
\textbf{Overview}
In this paper, we propose RA-MMEA, a reliability-aware framework for robust multi-modal entity alignment. As shown in Fig.~\ref{fig:model}, given an entity containing visual image, graph structure, relations, and attributes, we first extract their embedding (Sec.~\ref{embedding}). Afterwards, DA-VRP (Sec.~\ref{da-vrp}) is applied to predict the reliability of the image. If the image is reliable, we perform multi-modal fusion (Sec.~\ref{multi-modal fusion}) to obtain the entity representation; otherwise, we discard the unreliable visual embedding and apply SR-VEG (Sec.~\ref{sr-veg}) to produce a virtual visual embedding for modality fusion (Sec.~\ref{multi-modal fusion}).

\subsection{Multi-modal Knowledge Embedding} \label{embedding}
In this section, we describe embedding extraction for the visual image, graph structure, relations, and attributes of the entity $e_k$.

\vspace{0.2em}
\noindent
\textbf{Visual Embedding.} Given a visual image of the $k$-th entity $e_k$, we apply a pre-trained CLIP model~\cite{radford2021learning}, followed by fully connected (FC) layers, to extract the visual embedding $h_k^v$.

\vspace{0.2em}
\noindent
\textbf{Graph Structure Embedding.} For the graph structure of $e_k$, we follow~\cite{li2025probing,chen2023meaformer} and use graph attention network~\cite{velivckovic2018graph} with two attention heads and two layers to extract the graph embedding $h_k^g$.

\vspace{0.2em}
\noindent
\textbf{Relation Embedding.} Following~\cite{yang2019aligning, chen2023meaformer}, for the relations of $e_k$, we utilize bag-of-words followed by FC layers to extract the relation embedding $h_k^r$.

\vspace{0.2em}
\noindent
\textbf{Attribute Embedding.} For the attribute modality of $e_k$, we first employ bag-of-words and BERT~\cite{devlin2019bert} to generate two attribute embeddings, respectively. These embeddings are then combined through learnable weights and FC layers to obtain the final attribute embedding $h_k^a$.

\subsection{Dependency-Aware Visual Reliability Prediction (DA-VRP)}\label{da-vrp}

In RA-MMEA, we propose DA-VRP to estimate whether the visual modality of an entity is reliable under the global multi-modal context. The key intuition is that reliable images should exhibit coherent semantic dependency with the remaining modalities of an entity, and thus have relatively high dependency score with other modalities.

\vspace{0.3em}
\noindent
\textbf{Dependency Relation Modeling (DRM).} Specifically, we first perform dependency relation modeling (DRM) to compute the global relevance of each modality with the remaining modalities. More concretely, drawing inspiration from the Transformer architecture~\cite{vaswani2017attention}, we first apply the scalable dot-product to calculate the relevance score with normalization between modalities $m$ and $j$, as follows,
\begin{equation}
    s_{mj} = \frac{\exp(\phi(h_k^m)^{\top}  \phi(h_k^j)/d)}{\sum\nolimits_{n \in M} \exp(\phi(h_k^m)^{\top} \phi(h_k^n)/d)}
\end{equation}
where $s_{mj}$ is the relevance score between modality $m \in M$ and $j \in M$, and $M=\{v,g,r,a\}$ with $v$, $g$, $r$, $a$ representing the visual image, graph structure, relation, and attribute modalities. $d$ is a scaling factor, and $\phi(\cdot)$ the linear projection. 

Afterwards, we aggregate all relevance scores of each modality $m$ with normalization to compute its dependency score within the entity $e_k$, as follows,
\begin{equation} 
    w^m_k = \frac{\exp ( \sum\nolimits_{j \in M}s_{mj} )}{\sum\nolimits_{n \in M} \exp ( \sum\nolimits_{j \in M} s_{nj} )}
\end{equation}
where $w^m_k$ denotes the dependency score for modality $m$. After this, we employ the dependency score $w^v_k$ of visual image to predict the visual reliability.

\vspace{0.3em}
\noindent
\textbf{Visual Reliability Prediction.} The dependency score $w^k_v \in [0,1]$ reflects the semantic consistency between the visual modality and the remaining modalities of entity $e_k$, and we use it as a task-level proxy for visual reliability. This is motivated by the fact that cross-modal dependency provides a differentiable approximation of semantic consistency, a principled indicator of whether visual information agrees with the overall entity semantics. A higher dependency score indicates that the visual representation is more coherent with the overall entity semantics. In contrast, a lower score suggests that the image may be noisy, semantically inconsistent, or provide limited visual evidence for representing the corresponding entity. 

To predict the visual reliability, we first construct a probability vector based on $w^k_v$, as follows,
\begin{equation}
    p^v_{k} = [1 - w^v_k, w^v_k]
\end{equation}
where $p^v_{k,0}=1 - w^k_v$ denotes the probability that the image is unreliable, and $p^v_{k,1}=w^v_k$ represents the probability that the image is reliable. To obtain the binary reliability prediction, a straightforward way is to apply the Bernoulli function on $p^v_{k,0}$ and $p^v_{k,1}$. However, this operation is non-differentiable, preventing the entire pipeline from being trained in an end-to-end manner.

To handle this, we adopt Gumbel-Softmax~\cite{jang2017categorical,maddison2017concrete} to sample from the categorical distribution, as follows,
\begin{equation}
    g^{v}_k = \text{Gumbel-Softmax} \left(p^v_k\right)
\end{equation}
where $g^{v}_k$ denotes a binary one-hot vector representing the predicted visual reliability, with $g^{v}_k=[0,1]$ indicating that the image is reliable and $g^{v}_k=[1,0]$ indicating that the image is unreliable. By employing the Gumbel-Softmax trick, the entire framework becomes fully end-to-end trainable, thereby improving entity alignment performance.

\subsection{Stability-Regularized Visual Embedding Generation (SR-VEG)}\label{sr-veg}
To compensate for the loss of visual information caused by filtering the unreliable visual embedding, we propose SR-VEG to generate semantically consistent virtual visual embeddings from the remaining reliable modalities. Besides handling filtered visual modalities, it can also naturally generalize to entities that inherently lack visual information.

Specifically, we first construct a hybrid representation from the non-visual modalities, including graph structure, relation, and attribute embeddings, with the learned dependency scores from DA-VRP, as follows,
\begin{equation}\label{hybrid}
    h_k^{\text{hyb}} = \oplus_{m \in T} \left[ w_k^m \cdot h_k^m \right], \quad T = \{g, r, a \}
\end{equation}
Here, $\oplus$ denotes concatenation. By incorporating modality-aware weights, SR-VEG preserves cross-modal dependency during visual embedding generation. However, directly removing the visual modality may perturb the original multi-modal dependency structure. Although unreliable visual information should be suppressed, its removal can still affect dependency estimation among the remaining modalities, leading to unstable modality weights and potentially degrading the generated visual representation.

To alleviate this issue, we introduce a stability regularization term to constrain the discrepancy between dependency weights estimated before and after visual modality removal. For entities whose visual modalities are filtered, we first recompute dependency scores $\hat{w}_{k}^{m}$ ($m \in \{g, r, a\}$) with the embeddings $h_k^{g}$, $h_k^{r}$, and $h_k^{a}$ using DRM in DA-VRP, and then compare them with the original dependency weights $w_{k}^{g}$, $w_{k}^{r}$, and $w_{k}^{a}$. We retain only positive deviations using a ReLU operation via
\begin{equation}
    \mathcal{L}_p = \sum\nolimits_{m \in T} \max \left( \hat{w}_k^m - w_k^m, 0 \right)
\end{equation}
Here, the penalty $\mathcal{L}_p$ does not directly construct the hybrid representation for generation. Instead, it acts as an auxiliary stability constraint that regularizes dependency estimation under modality perturbation. By discouraging excessive dependency shifts after visual removal, SR-VEG maintains more stable cross-modal relationships and improves the generated visual embeddings.

After obtaining $h_k^{\text{hyb}}$, we use a VAE-style framework~\cite{KingmaW13auto,sohn2015learning} to generate  the pseudo-visual embeddings. Specifically, the encoder first maps the hybrid representation into the parameters of a latent Gaussian distribution via
\begin{equation}
    [\mu_k \oplus \log(\sigma_k)^2]
    =
    \Phi_{\text{Enc}}(h_k^{hyb})
\end{equation}
where $\Phi_{\text{Enc}}(\cdot)$ is the encoder implemented with MLP layers, and $\mu_k$ and $\sigma_k$ denote the mean and standard deviation of the latent distribution, respectively. Then, latent variables are sampled through the reparameterization trick, as follows,
\begin{equation}
    z_k = \mu_k + \epsilon \odot \sigma_k,
    \quad
    \epsilon \sim \mathcal{N}(\mathbf{0}, \mathbf{I})
\end{equation}
Afterwards, the sampled latent representation is decoded to generate pseudo-visual embeddings via
\begin{equation}
    \bar{h}_k^{v}
    =
    \Phi_{\text{Dec}}(z_k)
\end{equation}
where $\Phi_{\text{Dec}}(\cdot)$ is the decoder implemented with MLP layers, and $\bar{h}_k^v$ denotes the generated pseudo-visual embeddings. $\bar{h}_k^v$ is expected to preserve semantic consistency with the remaining modalities while compensating for the unreliable or missing visual information. The optimization objective of SR-VEG is detailed in Section~\ref{Training Pipeline and Optimization}.

\subsection{Multi-modal Feature Fusion}\label{multi-modal fusion}

In RA-MMEA, the final entity representation is obtained by fusing  embeddings from different modalities in a reliability-aware manner. For each entity, when its visual modality is predicted to be reliable, we directly incorporate the original visual embedding into the entity representation. Otherwise, we replace the unreliable visual embedding with the generated pseudo-visual embedding from SR-VEG. This process can be formulated as follows,
\begin{equation}
    h_k =
    w_k^g h_k^g
    \oplus
    w_k^r h_k^r
    \oplus
    w_k^a h_k^a
    \oplus
    \tilde{h}_k^v
\end{equation}
where $h_k$ denotes the final representation of entity $e_k$. $\tilde{h}_k^v =g_{k,1}^v w_k^v h_k^v+g_{k,0}^v \bar{h}_k^v$ is the reliability-aware visual representation, where $g_{k,1}^v$ and $g_{k,0}^v$ are the two elements of the visual reliability vector $g_{k}^v$, corresponding to the reliable and unreliable visual states, respectively.

\subsection{Optimization and Inference}
\label{Training Pipeline and Optimization} 
\textbf{Optimization.} To enhance interaction between modalities, we introduce multi-modal contrastive learning to achieve bidirectional alignment. The loss function is defined as follows,
\begin{equation}
    \mathcal{L}_{m} = -\log\left( p_m(e_u, e_t) + p_m(e_t, e_u)/{2} \right)
\end{equation}
where $p_m$ denotes the alignment probability between two entities $e_u$ and $e_t$. Furthermore, the intra-modal loss $\mathcal{L}_{\text{IAL}}$ and inter-modal loss $\mathcal{L}_{\text{ICL}}$ can be represented as follows,
\begin{equation}
    \mathcal{L}_{\text{IAL}} = \sum_{m \in M}{\mathcal{L}_{m}} \;\;\;\;\; \mathcal{L}_{\text{ICL}} = \sum_{m \in M}{\mathcal{\hat{L}}_{m}} 
\end{equation}
where $\mathcal{\hat{L}}_{m}$ is the post-fusion variant of $\mathcal{L}_{m}$. To alleviate imbalance and enforce constraints on reconstructed features, we jointly optimize the SR-VEG loss function $\mathcal{L}_{\text{SR}}$ together with the contrastive loss. Specifically, inspired by \citet{chen2023rethinking}, we employ two reconstruction objectives, $\mathcal{L}^{vis}_{\text{Re}}$ and $\mathcal{L}^{hyb}_{\text{Re}}$, to minimize the distance between the pseudo and true embeddings, thereby emphasizing the differences between different features during the autoencoding process. In addition, minimizing the Kullback-Leibler (KL) divergence $\mathcal{L}_{\text{KL}}$ encourages the latent space to approximate a Gaussian distribution. Then, we introduce a similarity distillation loss $\mathcal{L}_{\text{Sim}}$ to quantify the discrepancy between the similarity matrix of the hybrid embeddings and that of the virtual image features. Finally, the overall loss function can be expressed as follows,
\begin{equation}
    \mathcal{L} = \mathcal{L}_{\text{IAL}} + \mathcal{L}_{\text{ICL}} + \mathcal{L}_{\text{SR}} + \gamma \cdot \mathcal{L}_{p}
\end{equation}
where $\gamma$ controls the strength of $\mathcal{L}_{p}$, and $ \mathcal{L}_{\text{SR}} = \mathcal{L}_{\text{KL}} + \mathcal{L}^{vis}_{\text{Re}} + \mathcal{L}^{hyb}_{\text{Re}} + \mathcal{L}_{\text{Sim}} $.

\vspace{0.3em}
\noindent
\textbf{Inference.} We employ the cosine similarity (\text{Sim}) to quantify the alignment probability. The similarity matrix between the source entity set $E$ and the target entity set $E^{'}$ is denoted as $\text{Sim} \langle E, E^{'} \rangle$.

\section{Experiments}

\textbf{Datasets.} We validate the effectiveness and robustness of RA-MMEA on two cross-KG datasets, FB15K-DB15K/YG15K~\cite{liu2019mmkg}, with 20\%, 50\%, and 80\% reference entity alignments as seed pairs, and three bilingual DBP15K datasets (ZH/JA/FR-EN) \citep{sun2017cross} under the unsupervised setting. Appendix~\ref{sec:Datasets Statistics} depicts the statistics of multi-modal datasets. The details of the evaluation metrics are shown in Appendix~\ref{sec:Evaluation Metrics}.

\begin{table*}[!t]
  \centering
  \renewcommand\arraystretch{0.9}
  \resizebox{0.98\textwidth}{!}{
  \setlength{\tabcolsep}{10pt}
  \begin{tabular}{cllcccccc} 
  \toprule 
  \multirow{2}{*}{\textbf{Setting}} & \multirow{2}{*}{\textbf{Models}} & \multirow{2}{*}{\textbf{Venue}} & \multicolumn{3}{c}{\textbf{FB15K-DB15K}} & \multicolumn{3}{c}{\textbf{FB15K-YG15K}} \\ 
  \cmidrule(lr){4-6} \cmidrule(lr){7-9} 
  & & & \textbf{Hits@1$\uparrow$} & \textbf{Hits@10$\uparrow$} & \textbf{MRR$\uparrow$} & \textbf{Hits@1$\uparrow$} & \textbf{Hits@10$\uparrow$} & \textbf{MRR$\uparrow$} \\ 
  \toprule 
  \multirow{12}{*}{\rotatebox[origin=c]{90}{\textbf{Non-iterative}}}
  & MMEA \citep{chen2020mmea} & KSEM'20 & 0.265 & 0.541 & 0.357 & 0.234 & 0.480 & 0.317 \\
  & EVA \citep{liu2021visual} & AAAI'21 & 0.199 & 0.448 & 0.283 & 0.153 & 0.361 & 0.224 \\
  & MSNEA \citep{chen2022multi} & KDD'22 & 0.114 & 0.296 & 0.175 & 0.103 & 0.249 & 0.153 \\
  & MCLEA \citep{lin2022multi} & COLING'22 & 0.295 & 0.582 & 0.393 & 0.254 & 0.484 & 0.332 \\
  & MoAlign \citep{li2023multi} & EMNLP'23 & 0.318 & 0.564 & 0.409 & 0.296 & 0.525 & 0.378 \\
  & MEAformer \citep{chen2023meaformer} & ACM MM'23 & 0.417 & 0.715 & 0.518 & 0.327 & 0.595 & 0.417 \\
  & PMF \citep{huang2024progressively} & ACL'24 & 0.539 & - & 0.620 & 0.459 & - & 0.539 \\
  & RICEA \citep{li2025probing} & ACL'25 & 0.471 & 0.720 & 0.557 & 0.411 & 0.658 & 0.497 \\
  & SGMEA \citep{cheng2025sgmea} & COLING'25 & \underline{0.543} & \underline{0.777} & \underline{0.625} & \underline{0.587} & \underline{0.826} & \underline{0.670} \\
  & RULE \citep{li2026learning} & ICLR'26 & 0.444 & - & 0.538 & 0.389 & - & 0.470 \\
  & HUMEA \citep{xing2026modality} & AAAI'26 & 0.512 & 0.764 & 0.598 & 0.439 & 0.699 & 0.528 \\
  & \cellcolor{cyan!10}$\mathbf{\rm RA\text{-}MMEA\;(Ours)}$ & \cellcolor{cyan!10}-- & \cellcolor{cyan!10}$\mathbf{0.849}$ & \cellcolor{cyan!10}$\mathbf{0.948}$ & \cellcolor{cyan!10}$\mathbf{0.879}$ & \cellcolor{cyan!10}$\mathbf{0.787}$ & \cellcolor{cyan!10}$\mathbf{0.928}$ & \cellcolor{cyan!10}$\mathbf{0.835}$ \\
  \midrule
  \multirow{12}{*}{\rotatebox[origin=c]{90}{\textbf{Iterative}}}
  & EVA \citep{liu2021visual} & AAAI'21 & 0.231 & 0.488 & 0.318 & 0.188 & 0.403 & 0.260 \\
  & MSNEA \citep{chen2022multi} & KDD'22 & 0.149 & 0.392 & 0.232 & 0.138 & 0.346 & 0.210 \\
  & MCLEA \citep{lin2022multi} & COLING'22 & 0.395 & 0.656 & 0.487 & 0.322 & 0.546 & 0.400 \\
  & MEAformer \citep{chen2023meaformer} & ACM MM'23 & 0.578 & 0.812 & 0.661 & 0.444 & 0.692 & 0.529 \\
  & PMF \citep{huang2024progressively} & ACL'24 & 0.624 & - & 0.702 & 0.543 & - & 0.620 \\
  & IBMEA \citep{su2024ibmea} & ACM MM'24 & 0.631 & 0.813 & 0.697 & 0.521 & 0.708 & 0.584 \\
  & PCMEA \citep{wang2024pseudo} & AAAI'24 & 0.676 & 0.887 & 0.728 & 0.590 & 0.835 & 0.646 \\
  & RICEA \citep{li2025probing} & ACL'25 & 0.567 & 0.804 & 0.652 & 0.516 & 0.733 & 0.593 \\
  & SGMEA \citep{cheng2025sgmea} & COLING'25 & 0.661 & 0.847 & 0.729 & 0.750 & 0.901 & 0.805 \\
  & EIEA \citep{wang2025explicit} & KDD'25 & 0.769 & 0.869 & 0.804 & 0.830 & 0.915 & 0.860 \\
  & DMEA \citep{xing2026enhancing} & WWW'26 & \underline{0.804} & \underline{0.896} & \underline{0.849} & \underline{0.898} & \underline{0.957} & \underline{0.934} \\
  & \cellcolor{cyan!10}$\mathbf{\rm RA\text{-}MMEA\;(Ours)}$ & \cellcolor{cyan!10}-- & \cellcolor{cyan!10}$\mathbf{0.915}$ & \cellcolor{cyan!10}$\mathbf{0.976}$ & \cellcolor{cyan!10}$\mathbf{0.937}$ & \cellcolor{cyan!10}$\mathbf{0.899}$ & \cellcolor{cyan!10}$\mathbf{0.964}$ & \cellcolor{cyan!10}$\mathbf{0.936}$ \\
  \bottomrule 
  \end{tabular}
  }

  \caption{\raggedright
  Results on two cross-KG datasets using 20\% of the reference entity alignments for training. The best results are shown in \textbf{bold} and the second best results are \underline{underlined}.}
  \label{tab:mono20seeds}
\end{table*}

\vspace{0.2em}
\noindent
\textbf{Implementation Details.} RA-MMEA is implemented in PyTorch on a machine with NVIDIA RTX A6000 GPUs. RA-MMEA is trained for 500 epochs, and optionally further trained by 500 additional epochs with iterative training. We apply AdamW optimizer ($\beta_{1} = 0.9$, $\beta_{2} = 0.999$) with batch size of 3,500. The scaling factor $d$ is set to 17. The parameter $\gamma$ that controls $\mathcal{L}_p$ is set to 15.

\vspace{0.2em}
\noindent
\textbf{Compared Methods.} To rigorously evaluate our framework's effectiveness, we compare it with 16 methods that focus on multi-modal fusion optimization as in Sec.~\ref{Overall Results}. Notably, several recent MMEA methods, such as TMEA~\citep{chen2024tackling}, LGEA~\citep{lu2025breaking}, and CLAEA~\citep{ji2025capturing}, introduce external knowledge for MMEA. Since our goal is to design a reliability-aware framework that detects and mitigates unreliable visual modalities by exploiting inherent multi-modal dependency, rather than enhancing inputs with additional external information, we exclude these information-enhancement methods to ensure a fair comparison. For RULE~\citep{li2026learning}, we adopt the variant without MLLM during testing.

\subsection{State-of-the-art Comparison}
\label{Overall Results}

Tab.~\ref{tab:mono20seeds} shows the non-iterative and iterative results on the cross-KG datasets with 20\% seed alignments. RA-MMEA consistently outperforms all baselines on both datasets and under both training settings across all metrics. Specifically, on FB15K-DB15K, RA-MMEA improves Hits@1 by 11.1\%--30.6\% and MRR by 8.8\%--25.4\% over the second-best baselines. On FB15K-YG15K, it achieves gains of 0.1\%--20.0\% in Hits@1 and 0.2\%--16.5\% in MRR. Notably, the improvements are particularly substantial under the non-iterative setting, demonstrating RA-MMEA's effectiveness even with limited seed alignments. Additional results with 50\% and 80\% seed alignments are provided in Appendix~\ref{sec:More Results}, while results on the bilingual DBP15K datasets are reported in Appendix~\ref{sec:Unsupervised Training}. Overall, these results demonstrate that RA-MMEA consistently achieves state-of-the-art performance across datasets, seed ratios, and training settings.

\subsection{Robustness Evaluation}

Visual information in real-world scenarios often contains varying degrees of noise, making it important for MMEA models to maintain stable alignment performance as visual information becomes increasingly unreliable. To evaluate this robustness, we construct artificial Entity-Image mismatches on FB15K-DB15K by randomly selecting 20\%, 50\%, and 80\% of entities and shuffling their associated images while preventing them from being reassigned to their original entities, thereby simulating progressively increasing visual noise.

\begin{figure}[!t]
  \centering
  \includegraphics[width=0.49\linewidth]{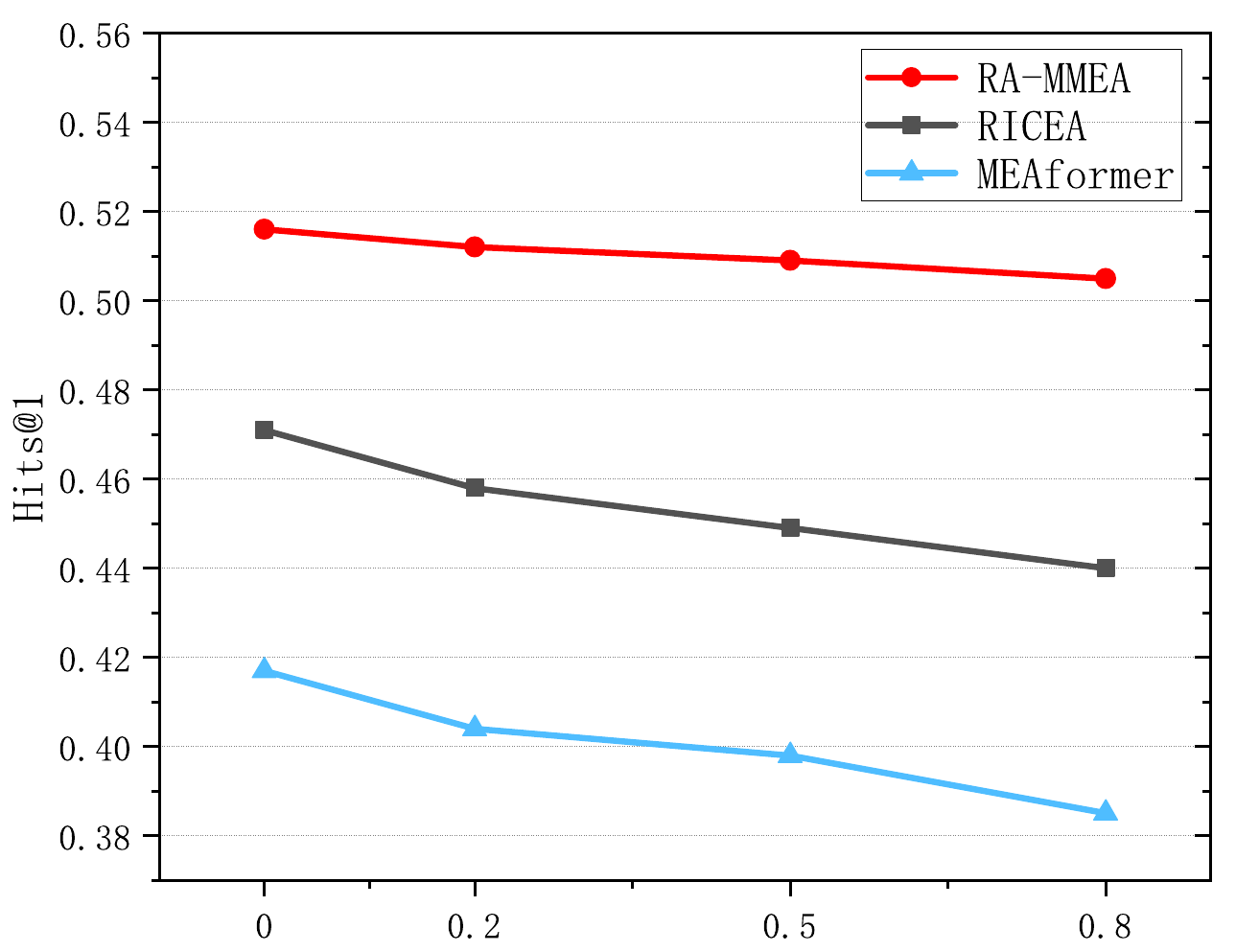}
  \includegraphics[width=0.49\linewidth]{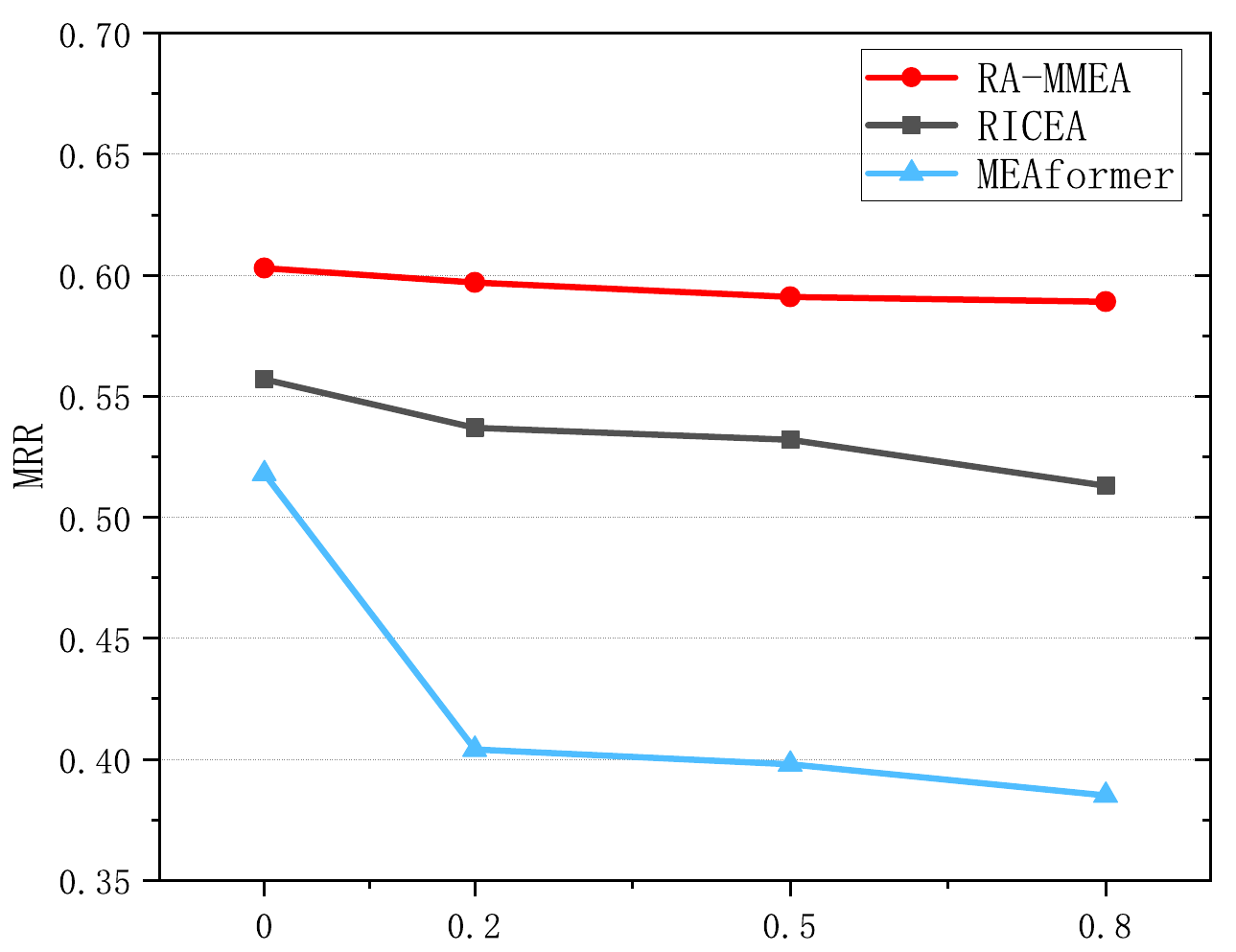}
  \captionsetup{justification=justified, singlelinecheck=false}
  \caption{Performance of strong baselines across visual noise levels on FB15K-DB15K.}
  \label{fig:robust}
\end{figure}

As shown in Fig.~\ref{fig:robust}, the performance of all methods gradually decreases as the noise ratio increases, indicating that incorrect Entity-Image associations make MMEA increasingly challenging. Nevertheless, RA-MMEA consistently achieves the best performance across all noise ratios and exhibits substantially smaller performance degradation than MEAformer and RICEA. These results demonstrate that DA-VRP effectively mitigates interference caused by incorrect Entity-Image associations, enabling RA-MMEA to preserve robust multi-modal representations under increasing visual noise and confirming its robustness in challenging noisy MMEA scenarios.

\subsection{Visual Noise Detection}
\label{Visual Noise Detection}

The above results demonstrate that RA-MMEA remains robust under progressively increasing visual noise. We further investigate whether this advantage stems from the ability of DA-VRP to identify unreliable visual information. Specifically, we treat the artificially injected Entity-Image mismatches as known unreliable samples and measure the proportion of these samples correctly identified as unreliable by DA-VRP. 

\begin{table}[!t]
  \centering
  \renewcommand\arraystretch{1.00}
  \setlength{\tabcolsep}{5pt}
  \resizebox{\columnwidth}{!}{
  \begin{tabular}{cccc}
    \toprule
    \textbf{Noise Ratio} & Ratio=20\% & Ratio=50\% & Ratio=80\% \\
    \midrule
    \textbf{Detection Rate (\%)} & 75.24 & 74.87 & 75.08 \\
    \bottomrule
  \end{tabular}
  }
  \captionsetup{justification=justified, singlelinecheck=false}
  \caption{Detection rate of DA-VRP under varying visual noise ratios.}
  \label{tab:detection}
\end{table}

As shown in Tab.~\ref{tab:detection}, DA-VRP achieves detection rates of 75.24\%, 74.87\%, and 75.08\% under visual noise ratios of 20\%, 50\%, and 80\%, respectively, remaining stable at approximately 75\% across different noise levels. These results indicate that DA-VRP can effectively identify unreliable visual information based on cross-modal semantic inconsistency, while its detection capability remains largely insensitive to changes in the proportion of injected noise. Since existing MMEA datasets do not provide ground-truth reliability annotations for the original Entity-Image pairs, we evaluate only the artificially injected visual noise with known labels. Nevertheless, the consistent detection performance provides empirical evidence that cross-modal dependency serves as an effective task-level proxy for visual reliability. Further qualitative evidence is provided in Appendix~\ref{Appendix: Case Study}, where the predicted reliability is consistent with the semantic correspondence between entities and their associated images.

\begin{table}[!t]
  \centering
  \renewcommand\arraystretch{1.00}
  \setlength{\tabcolsep}{5pt}
  \resizebox{\linewidth}{!}{
  \begin{tabular}{cccccc}
    \hline
     & \textbf{DA-VRP} & \textbf{SR-VEG} & \textbf{Hits@1$\uparrow$} & \textbf{Hits@10$\uparrow$} & \textbf{MRR$\uparrow$} \\
    \hline
    \ding{182} & \ding{55} & \ding{55} & 0.682 & 0.860 & 0.733 \\
    \ding{183} & \ding{51} & \ding{55} & 0.798 & 0.886 & 0.855 \\
    \ding{184} & \ding{55} & \ding{51} & 0.701 & 0.873 & 0.752 \\
    \rowcolor{cyan!10}
    \ding{185} & \ding{51} & \ding{51} & 0.849 & 0.948 & 0.879 \\
    \hline
  \end{tabular}
  }
  \captionsetup{justification=justified, singlelinecheck=false}
  \caption{Ablation studies on DA-VRP and SR-VEG.}
  \label{tab:component}
\end{table}

\begin{table}[!t]
  \centering
  \renewcommand\arraystretch{1.00}
  \setlength{\tabcolsep}{7pt}
  \resizebox{\linewidth}{!}{
  \begin{tabular}{ccccc}
    \hline
    &  & \textbf{Hits@1$\uparrow$} & \textbf{Hits@10$\uparrow$} & \textbf{MRR$\uparrow$} \\
    \hline
    \ding{182} & DA-VRP w/ BS & 0.503 & 0.762 & 0.590 \\
      \rowcolor{cyan!10}
    \ding{183} & DA-VRP w/ GS & 0.849 & 0.948 & 0.879 \\
    \hline
  \end{tabular}
  }
  \captionsetup{justification=justified, singlelinecheck=false}
  \caption{Ablation studies on Gumbel-Softmax (GS) and Bernoulli Sampling (BS) in DA-VRP.}
  \label{tab:bd}
\end{table}

\begin{table}[!t]
  \centering
  \renewcommand\arraystretch{1.00}
  \setlength{\tabcolsep}{7pt}
  \resizebox{\linewidth}{!}{
  \begin{tabular}{ccccc}
    \hline
    &  & \textbf{Hits@1$\uparrow$} & \textbf{Hits@10$\uparrow$} & \textbf{MRR$\uparrow$} \\
    \hline
    \ding{182} & SR-VEG w/o $\mathcal{L}_p$ & 0.833 & 0.941 & 0.874 \\
    \rowcolor{cyan!10}
    \ding{183} & SR-VEG w/ $\mathcal{L}_p$ & 0.849 & 0.948 & 0.879 \\
    \hline
  \end{tabular}
  }
  \captionsetup{justification=justified, singlelinecheck=false}
  \caption{Ablation studies on $\mathcal{L}_p$ for SR-VEG.}
  \label{tab:penalty}
\end{table}

\begin{table}[!t]
  \centering
  \renewcommand\arraystretch{1.00}
  \setlength{\tabcolsep}{7pt}
  \resizebox{\linewidth}{!}{
  \begin{tabular}{ccccc}
    \hline
     & & \textbf{Hits@1$\uparrow$} & \textbf{Hits@10$\uparrow$} & \textbf{MRR$\uparrow$} \\
    \hline
    \ding{182} & generation w/ $\hat{w}_k^m$ & 0.830 & 0.938 & 0.867 \\
    \rowcolor{cyan!10}
    \ding{183} & generation w/ $w_k^m$ & 0.849 & 0.948 & 0.879 \\
    \hline
  \end{tabular}
  }
  \captionsetup{justification=justified, singlelinecheck=false}
  \caption{Ablation studies on using $\hat{w}_k^m$ and using $w_k^m$ for hybrid representation in SR-VEG.}
  \label{tab:weight}
\end{table}

\begin{table}[!t]
  \centering
  \renewcommand\arraystretch{1.00}
  \setlength{\tabcolsep}{12pt}
  \resizebox{1\linewidth}{!}{
  \begin{tabular}{ccccc}
    \hline
    & & \textbf{Hits@1$\uparrow$} & \textbf{Hits@10$\uparrow$} & \textbf{MRR$\uparrow$} \\
    \hline
    \ding{182} & $\gamma=1$ & 0.834 & 0.942 & 0.874 \\
    \ding{183} & $\gamma=5$ & 0.836 & 0.943 & 0.875 \\
    \ding{184} &$\gamma=10$ & 0.845 & 0.946 & 0.877 \\
    \rowcolor{cyan!10}
    \ding{185} &$\gamma=15$ & 0.849 & 0.948 & 0.879 \\
    \ding{186} &$\gamma=20$ & 0.840 & 0.945 & 0.879 \\
    \hline
  \end{tabular}
  }
  \captionsetup{justification=justified, singlelinecheck=false}
  \caption{Ablation studies on the parameter $\gamma$ for $\mathcal{L}_p$.}
  \label{tab:gamma}
\end{table}

\subsection{Ablation Study}

To further analyze RA-MMEA, we conduct ablation studies on FB15K-DB15K with 20\% seeds. Our final configuration is highlighted in \textcolor{cyan}{cyan}.

\vspace{0.3em}
\noindent
\textbf{Impacts of DA‑VRP and SR‑VEG.} The core of RA-MMEA lies in two key modules, including DA-VRP and SR-VEG. To analyze their impacts, we conduct ablation studies in Tab.~\ref{tab:component}. As shown, without DA-VRP and SR-VEG, the Hits@1 score is 0.682 (\ding{182}). When using DA‑VRP alone, the Hits@1 score is largely improved to 0.798 with 11.6\% gains (\ding{183}); when applying SR‑VEG only, the Hits@1 score is boosted to 0.701 with moderate improvements (\ding{184}). Using both modules together yields the best performance of 0.849 Hits@1 score with 16.7\% gains (\ding{185}), confirming their strong complementarity for improving MMEA.

\vspace{0.3em}
\noindent
\textbf{Gumbel‑Softmax \emph{v.s.} Bernoulli Sampling.} In our DA-VRP, we adopt Gumbel-Softmax to predict the visual reliability, enabling end-to-end training of RA-MMEA. To analyze its effectiveness, we conduct ablation studies by comparing Gumbel-Softmax and Bernoulli sampling for visual reliability prediction. The results are reported in Tab.~\ref{tab:bd}. As shown, we can observe that, when using Bernoulli sampling for reliability prediction, the Hits@1 score is 0.503 (\ding{182}). In contrast,  when applying Gumbel-Softmax, the Hits@1 score is improved to 0.849 with 34.6\% gains (\ding{183}). This confirms that differentiable reliability estimation is essential for learning accurate modality‑filtering behavior.

\vspace{0.3em}
\noindent
\textbf{Impact of $\mathcal{L}_p$.} In SR-VEG, we design a stability regularization term $\mathcal{L}_p$, which aims to constrain dependency shifts after removing the visual modality. To analyze the impact of $\mathcal{L}_p$, we conduct the ablation studies in Tab.~\ref{tab:penalty}. As shown, without $\mathcal{L}_p$, the model suffers from degraded dependency estimation, leading to suboptimal pseudo‑visual embeddings and performance. Incorporating $\mathcal{L}_p$ consistently improves the Hits@1 score from 0.833 (\ding{182}) to 0.849 (\ding{183}) with 1.6\% gains, demonstrating that stabilizing cross‑modal dependency is critical for reliable virtual visual embedding generation.

\vspace{0.3em}
\noindent
\textbf{Analysis of $w_k^m$ for Hybrid Representation.} In SR-VEG, we apply the original dependency scores $w_k^m$ ($m \in \{g,r,a\}$) to construct the hybrid representation in Eq.~(\ref{hybrid}). To validate its effectiveness, we compare hybrid representation construction with $w_k^m$ and $\hat{w}_k^m$ in Tab.~\ref{tab:weight}. As shown, using the original dependency scores $w_k^m$ is more stable (\ding{183} \emph{v.s.} \ding{182}), as they are computed with full multi-modal information, whereas re‑estimating weights after removing the visual modality relies on less information and therefore becomes unstable.

\vspace{0.3em}
\noindent
\textbf{Impact of $\gamma$.} To analyze the impact of $\gamma$ for $\mathcal{L}_p$, we conduct ablation studies in Tab.~\ref{tab:gamma} by changing the value of $\gamma$. As shown in Tab.~\ref{tab:gamma}, we can see that, when $\gamma = 15$, we show the best performance (\ding{185}). Therefore, we empirically set $\gamma$ to 15.

\subsection{Plug-and-Play Capability}

We integrate DA-VRP and SR-VEG into two MMEA frameworks with similar processing pipelines, MEAformer and RICEA, to evaluate the generalizability and plug-and-play capability of the proposed modules. As shown in Tab.~\ref{tab:Plug}, incorporating DA-VRP and SR-VEG consistently improves both frameworks across all evaluation metrics. Specifically, MEAformer achieves improvements of 8.9\% and 7.6\% in Hits@1 and MRR, respectively, while RICEA obtains corresponding gains of 3.0\% and 3.2\%. These consistent improvements demonstrate that DA-VRP and SR-VEG are not restricted to a specific MMEA architecture but can be flexibly integrated into existing frameworks to enhance their ability to effectively handle diverse unreliable visual information.

\begin{table}[!t]
  \centering
  \renewcommand\arraystretch{1.1}
  \setlength{\tabcolsep}{0.5pt}
  \resizebox{\linewidth}{!}{
  \begin{tabular}{lccc} 
  \hline
  \textbf{Models} & \textbf{\tabincell{c}{Hits@1}$\uparrow$} & \textbf{\tabincell{c}{Hits@10}$\uparrow$} & \textbf{\tabincell{c}{MRR}$\uparrow$} \\ 
  \hline
  MEAformer~\citep{chen2023meaformer} & 0.417 & 0.715 & 0.518 \\
  MEAformer\;(+) & 0.506 & 0.758  & 0.594 \\
  \hline
  \textbf{Improvement\;(\%)} & \textbf{8.9} & \textbf{4.3} & \textbf{7.6} \\
  \hline
  RICEA~\cite{li2025probing} & 0.471 & 0.720 & 0.557 \\
  RICEA\;(+) & 0.501 & 0.755 & 0.589 \\
  \hline
  \textbf{Improvement\;(\%)} & \textbf{3.0} & \textbf{3.5} & \textbf{3.2} \\
  \hline 
  \end{tabular}
  }
  \captionsetup{justification=justified, singlelinecheck=false}
  \caption{Plug-and-play evaluation of DA-VRP and SR-VEG across representative MMEA frameworks. "+" denotes their variants.}
  \label{tab:Plug}
\end{table}

% Due to space limitation, we provide more analyses and experimental results in the Appendices.

\section{Conclusion}

In this paper, we propose RA-MMEA, a reliability-aware framework for robust multi-modal entity alignment. RA-MMEA estimates visual reliability through cross-modal dependency and generates alternative representations for unreliable images. Extensive experiments demonstrate its effectiveness and robustness, highlighting the importance of reliability-aware visual modeling in MMEA.

\newpage
\section*{Limitations}

Although RA-MMEA demonstrates strong performance across multiple benchmarks, existing MMEA datasets typically provide only limited and relatively static visual information for each entity, which may not fully capture the diversity of entity-related images in real-world scenarios. In practice, the same entity may be associated with images exhibiting different visual content across different contexts, making it challenging to evaluate MMEA models under more diverse visual conditions. 

Exploring more diverse visual settings could further broaden the applicability of reliability-aware MMEA. It would also provide additional insights into visual reliability across different entity types and scenarios. Such investigations could further enrich our understanding of reliability-aware visual modeling in MMEA. These directions may also offer further insights into how visual information contributes to multi-modal entity alignment. Future work could develop MMEA benchmarks with richer and more diverse visual information, enabling a more comprehensive evaluation of the robustness and generalizability of MMEA methods in real-world scenarios.

% \section*{Acknowledgments}


\bibliography{custom}

\newpage
\appendix

\section{Appendix: Datasets Statistics}
\label{sec:Datasets Statistics}

Tab.~\ref{tab:Datasets} shows the statistics of datasets, including the number of entities (Ent.), relations (Rel.), attributes (Attr.), number of relation triples (Rel Tr.) and attribute triples (Attr Tr.), number of images (Image), and number of the reference entity pairs (EA pairs). Each entity is associated to multiple attributes and 0 $\sim$ 1 image.

\begin{table*}[hbp]
  \centering  
  \renewcommand\arraystretch{0.90}
  \setlength{\tabcolsep}{12pt}
  \resizebox{\textwidth}{!}{
  \begin{tabular}{@{}c|c|c|c|c|c|c|c|c@{}} 
  \toprule 
  Dataset & KG & Ent. & Rel. & Attr. & Rel. Tr. & Attr. Tr. & Image & EA pairs \\
  \toprule
  \multirow{2}{*}{FB15K-DB15K} & FB15K & 14,951 & 1,345 & 116 & 592,213 & 29,395 & 14,836 & \multirow{2}{*}{12,846} \\
   & DB15K & 12,842 & 279 & 225 & 89,197 & 48,080 & 12,812 &  \\
  \toprule
  \multirow{2}{*}{FB15K-YG15K} & FB15K & 14,951 & 1,345 & 116 & 592,213 & 29,395 & 14,836 & \multirow{2}{*}{11,199} \\
  & YG15K & 15,404 & 32 & 7 & 122,886 & 23,532 & 11,172 &  \\
  \toprule
  \multirow{2}{*}{DBP15K\(_{ZH-EN}\)} & ZH & 19,388 & 1,701 & 8,111 & 70,414 & 248,035 & 15,912 & \multirow{2}{*}{15,000} \\
  & EN & 19,572 & 1,323 & 7,173 & 95,142 & 343,218 & 14,125 &  \\
  \toprule
  \multirow{2}{*}{DBP15K\(_{JA-EN}\)} & JA & 19,814 & 1,299 & 5,882 & 77,214 & 248,991 & 12,739 & \multirow{2}{*}{15,000} \\
  & EN & 19,780 & 1,153 & 6,066 & 93,484 & 320,616 & 13,741 &  \\
  \toprule
  \multirow{2}{*}{DBP15K\(_{FR-EN}\)} & FR & 19,661 & 903 & 4,547 & 105,998 & 273,825 & 14,174 & \multirow{2}{*}{15,000} \\
  & EN & 19,993 & 1,208 & 6,422 & 115,722 & 351,094 & 13,858 &  \\
  \bottomrule 
  \end{tabular}
  }
  \caption{\raggedright Statistics of MMEA datasets, with EA pairs representing the reference entity alignments.
  }
  \label{tab:Datasets}
\end{table*}

\section{Appendix: Evaluation Metrics}
\label{sec:Evaluation Metrics}

We adopt Hits@n and MRR as evaluation metrics for all methods. Hits@n measures the proportion of correct entities ranked within the top n according to similarity scores, while MRR represents the mean reciprocal rank of the correct entities. Higher values of Hits@n and MRR indicate superior performance of the evaluated method.

\section{Appendix: More Results on Cross-KG Datasets}
\label{sec:More Results}

Tabs.~\ref{tab:mono50seeds} and~\ref{tab:mono80seeds} report the results on the cross-KG datasets using 50\% and 80\% of the reference entity alignments as seed pairs, respectively. RA-MMEA consistently achieves the best performance across all settings. Specifically, with 50\% seed alignments, RA-MMEA improves Hits@1 by 9.2\%--20.0\% and MRR by 7.2\%--16.2\% over the second-best baselines. With 80\% seed alignments, it further achieves gains of 6.0\%--13.6\% in Hits@1 and 4.4\%--11.1\% in MRR.

As the proportion of seed alignments increases, the performance of most methods improves due to the stronger alignment supervision. Nevertheless, RA-MMEA maintains a clear advantage over existing methods under both non-iterative and iterative settings. Notably, the improvements remain substantial even with 80\% seed alignments, where competing methods already achieve strong alignment performance. This indicates that the benefit of reliability-aware visual modeling is not limited to low-supervision scenarios but remains effective when abundant alignment supervision is available.

Together with the results using 20\% seed alignments in Tab.~\ref{tab:mono20seeds}, these results demonstrate that RA-MMEA consistently benefits from reliable visual information across different levels of alignment supervision, further confirming the effectiveness and generalizability of the proposed framework on cross-KG datasets.

\begin{table*}[!t]
  \centering
  \renewcommand\arraystretch{0.86}
  \resizebox{0.98\textwidth}{!}{
  \setlength{\tabcolsep}{12pt}
  \begin{tabular}{clcccccc} 
  \toprule 
  \multirow{2}{*}{\textbf{Setting}} & \multirow{2}{*}{\textbf{Models}} & \multicolumn{3}{c}{\textbf{FB15K-DB15K}} & \multicolumn{3}{c}{\textbf{FB15K-YG15K}} \\ 
  \cmidrule(lr){3-5} \cmidrule(lr){6-8} 
  & & \textbf{Hits@1$\uparrow$} & \textbf{Hits@10$\uparrow$} & \textbf{MRR$\uparrow$} & \textbf{Hits@1$\uparrow$} & \textbf{Hits@10$\uparrow$} & \textbf{MRR$\uparrow$} \\ 
  \toprule 
  \multirow{10}{*}{\rotatebox[origin=c]{90}{\textbf{Non-iterative}}}
  & \multirow{1}*{MMEA \citep{chen2020mmea}} & 0.417 & 0.703 & 0.512 & 0.403 & 0.645 & 0.486 \\ 
  & \multirow{1}*{EVA \citep{liu2021visual}} & 0.334 & 0.589 & 0.422 & 0.311 & 0.534 & 0.388 \\
  & \multirow{1}*{MSNEA \citep{chen2022multi}} & 0.288 & 0.590 & 0.388 & 0.320 & 0.589 & 0.413 \\
  & \multirow{1}*{MCLEA \citep{lin2022multi}} & 0.555 & 0.784 & 0.637 & 0.501 & 0.705 & 0.574 \\
  & \multirow{1}*{MoAlign \citep{li2023multi}} & 0.576 & 0.749 & 0.634 & 0.550 & 0.713 & 0.617 \\
  & \multirow{1}*{MEAformer \citep{chen2023meaformer}} & 0.619 & 0.843 & 0.698 & 0.560 & 0.778 & 0.639 \\
  & \multirow{1}*{RICEA \citep{li2025probing}} & 0.648 & 0.852 & 0.721 & 0.617 & 0.811 & 0.687 \\
  & \multirow{1}*{SGMEA \citep{cheng2025sgmea}} & \underline{0.716} & \underline{0.882} & \underline{0.775} & \underline{0.780} & \underline{0.924} & \underline{0.832} \\
  & \multirow{1}*{HUMEA \citep{xing2026modality}} & 0.695 & 0.880 & 0.763 & 0.649 & 0.847 & 0.720 \\
  & \cellcolor{cyan!10}\multirow{1}*{$\mathbf{\rm RA\text{-}MMEA\;(Ours)}$} & \cellcolor{cyan!10}$\mathbf{0.916}$ & \cellcolor{cyan!10}$\mathbf{0.978}$ & \cellcolor{cyan!10}$\mathbf{0.937}$ & \cellcolor{cyan!10}$\mathbf{0.888}$ & \cellcolor{cyan!10}$\mathbf{0.969}$ & \cellcolor{cyan!10}$\mathbf{0.917}$ \\
  \toprule 
  \multirow{9}{*}{\rotatebox[origin=c]{90}{\textbf{Iterative}}}
  & \multirow{1}*{EVA \citep{liu2021visual}} & 0.364 & 0.606 & 0.449 & 0.325 & 0.560 & 0.404 \\
  & \multirow{1}*{MSNEA \citep{chen2022multi}} & 0.358 & 0.656 & 0.459 & 0.376 & 0.646 & 0.472 \\
  & \multirow{1}*{MCLEA \citep{lin2022multi}} & 0.620 & 0.832 & 0.696 & 0.563 & 0.751 & 0.631 \\
  & \multirow{1}*{MEAformer \citep{chen2023meaformer}} & 0.690 & 0.871 & 0.755 & 0.612 & 0.808 & 0.682 \\
  & \multirow{1}*{IBMEA \citep{su2024ibmea}} & 0.742 & 0.880 & 0.793 & 0.655 & 0.821 & 0.714 \\
  & \multirow{1}*{PCMEA \citep{wang2024pseudo}} & 0.738 & \underline{0.915} & 0.781 & 0.670 & 0.886 & 0.721 \\
  & \multirow{1}*{RICEA \citep{li2025probing}} & 0.692 & 0.869 & 0.757 & 0.658 & 0.827 & 0.720 \\
  & \multirow{1}*{SGMEA \citep{cheng2025sgmea}} & \underline{0.752} & 0.894 & \underline{0.802} & \underline{0.827} & \underline{0.938} & \underline{0.868} \\
  & \cellcolor{cyan!10}\multirow{1}*{$\mathbf{\rm RA\text{-}MMEA\;(Ours)}$} & \cellcolor{cyan!10}$\mathbf{0.932}$ & \cellcolor{cyan!10}$\mathbf{0.984}$ & \cellcolor{cyan!10}$\mathbf{0.950}$ & \cellcolor{cyan!10}$\mathbf{0.919}$ & \cellcolor{cyan!10}$\mathbf{0.977}$ & \cellcolor{cyan!10}$\mathbf{0.940}$ \\
  \bottomrule 
  \end{tabular}
  }
  \caption{\raggedright Results on two cross-KG datasets using 50\% of the reference entity alignments for training. The best results are shown in \textbf{bold} and the second best results are \underline{underlined}.
  }
  \label{tab:mono50seeds}
\end{table*}

\begin{table*}[!t]
  \centering
  \renewcommand\arraystretch{0.86}
  \resizebox{0.98\textwidth}{!}{
  \setlength{\tabcolsep}{12pt}
  \begin{tabular}{clcccccc} 
  \toprule 
  \multirow{2}{*}{\textbf{Setting}} & \multirow{2}{*}{\textbf{Models}} & \multicolumn{3}{c}{\textbf{FB15K-DB15K}} & \multicolumn{3}{c}{\textbf{FB15K-YG15K}} \\ 
  \cmidrule(lr){3-5} \cmidrule(lr){6-8} 
  & & \textbf{Hits@1$\uparrow$} & \textbf{Hits@10$\uparrow$} & \textbf{MRR$\uparrow$} & \textbf{Hits@1$\uparrow$} & \textbf{Hits@10$\uparrow$} & \textbf{MRR$\uparrow$} \\ 
  \toprule 
  \multirow{10}{*}{\rotatebox[origin=c]{90}{\textbf{Non-iterative}}}
  & \multirow{1}*{MMEA \citep{chen2020mmea}} & 0.590 & 0.869 & 0.685 & 0.598 & 0.839 & 0.682 \\
  & \multirow{1}*{EVA \citep{liu2021visual}} & 0.484 & 0.696 & 0.563 & 0.491 & 0.692 & 0.565 \\
  & \multirow{1}*{MSNEA \citep{chen2022multi}} & 0.518 & 0.779 & 0.613 & 0.531 & 0.778 & 0.620 \\
  & \multirow{1}*{MCLEA \citep{lin2022multi}} & 0.735 & 0.890 & 0.790 & 0.667 & 0.824 & 0.722 \\
  & \multirow{1}*{MoAlign \citep{li2023multi}} & 0.699 & 0.882 & 0.773 & 0.689 & 0.884 & 0.769 \\
  & \multirow{1}*{MEAformer \citep{chen2023meaformer}} & 0.765 & 0.916 & 0.820 & 0.703 & 0.873 & 0.766 \\
  & \multirow{1}*{RICEA \citep{li2025probing}} & 0.776 & 0.916 & 0.829 & 0.734 & 0.892 & 0.792 \\
  & \multirow{1}*{SGMEA \citep{cheng2025sgmea}} & \underline{0.815} & 0.931 & 0.828 & \underline{0.857} & \underline{0.951} & \underline{0.894} \\
  & \multirow{1}*{HUMEA \citep{xing2026modality}} & 0.809 & \underline{0.935} & \underline{0.856} & 0.774 & 0.920 & 0.827 \\
  & \cellcolor{cyan!10}\multirow{1}*{$\mathbf{\rm RA\text{-}MMEA\;(Ours)}$} & \cellcolor{cyan!10}$\mathbf{0.951}$ & \cellcolor{cyan!10}$\mathbf{0.990}$ & \cellcolor{cyan!10}$\mathbf{0.967}$ & \cellcolor{cyan!10}$\mathbf{0.939}$ & \cellcolor{cyan!10}$\mathbf{0.987}$ & \cellcolor{cyan!10}$\mathbf{0.957}$ \\
  \toprule 
  \multirow{9}{*}{\rotatebox[origin=c]{90}{\textbf{Iterative}}}
  & \multirow{1}*{EVA \citep{liu2021visual}} & 0.491 & 0.711 & 0.573 & 0.493 & 0.695 & 0.572 \\
  & \multirow{1}*{MSNEA \citep{chen2022multi}} & 0.565 & 0.810 & 0.651 & 0.593 & 0.806 & 0.668 \\
  & \multirow{1}*{MCLEA \citep{lin2022multi}} & 0.741 & 0.900 & 0.802 & 0.681 & 0.837 & 0.737 \\
  & \multirow{1}*{MEAformer \citep{chen2023meaformer}} & 0.784 & 0.921 & 0.834 & 0.724 & 0.880 & 0.783 \\
  & \multirow{1}*{IBMEA \citep{su2024ibmea}} & 0.821 & 0.922 & 0.859 & 0.751 & 0.890 & 0.800 \\
  & \multirow{1}*{PCMEA \citep{wang2024pseudo}} & 0.820 & \underline{0.964} & 0.858 & 0.756 & 0.942 & 0.802 \\
  & \multirow{1}*{RICEA \citep{li2025probing}} & 0.787 & 0.919 & 0.838 & 0.752 & 0.899 & 0.804 \\
  & \multirow{1}*{SGMEA \citep{cheng2025sgmea}} & \underline{0.828} & 0.921 & \underline{0.861} & \underline{0.882} & \underline{0.967} & \underline{0.915} \\
  & \cellcolor{cyan!10}\multirow{1}*{$\mathbf{\rm RA\text{-}MMEA\;(Ours)}$} & \cellcolor{cyan!10}$\mathbf{0.958}$ & \cellcolor{cyan!10}$\mathbf{0.991}$ & \cellcolor{cyan!10}$\mathbf{0.970}$ & \cellcolor{cyan!10}$\mathbf{0.942}$ & \cellcolor{cyan!10}$\mathbf{0.989}$ & \cellcolor{cyan!10}$\mathbf{0.959}$ \\
  \bottomrule 
  \end{tabular}
  }
  \caption{\raggedright Results on two cross-KG datasets using 80\% of the reference entity alignments for training. The best results are shown in \textbf{bold} and the second best results are \underline{underlined}.
  }
  \label{tab:mono80seeds}
\end{table*}

\section{Appendix: Unsupervised Training on Bilingual Datasets}
\label{sec:Unsupervised Training}

To validate the performance of our framework on the bilingual domain, we also conducted unsupervised experiments on the bilingual datasets DBP15K \citep{sun2017cross}. Unsupervised training was initially introduced by \citet{liu2021visual} to mitigate dependence on gold labels. Following \citet{lin2022multi}, the vision encoders $Enc_{v}$ are configured as ResNet-152 \citep{he2016deep} with a vision feature dimension of $d_{v} = 2048$. We utilize pre-trained 300-d GloVe vectors along with character bigrams for surface representation after applying machine translations for entity names.

\begin{table*}[!t]
  \centering
  \renewcommand\arraystretch{1.00}
  \resizebox{\textwidth}{!}{
  \begin{tabular}{lccccccccc} 
  \toprule 
  \multirow{2}{*}{Models} & \multicolumn{3}{c}{\textbf{$ \textbf{DBP15K}_{\emph{ZH-EN}} $}} & \multicolumn{3}{c}{\textbf{$ \textbf{DBP15K}_{\emph{JA-EN}} $}} & \multicolumn{3}{c}{\textbf{$ \textbf{DBP15K}_{\emph{FR-EN}} $}} \\ 
  \cmidrule(lr){2-4} \cmidrule(lr){5-7} \cmidrule(lr){8-10} 
  & \textbf{Hits@1$\uparrow$} & \textbf{Hits@10$\uparrow$} & \textbf{MRR$\uparrow$} & \textbf{Hits@1$\uparrow$} & \textbf{Hits@10$\uparrow$} & \textbf{MRR$\uparrow$} & \textbf{Hits@1$\uparrow$} & \textbf{Hits@10$\uparrow$} & \textbf{MRR$\uparrow$} \\ 
  \toprule
  \multicolumn{10}{c}{$ \textit{\textbf{Non-iterative}} $}\\
  \toprule  
  \multirow{1}*{EVA \citep{liu2021visual}} & 0.891 & 0.961 & 0.917 & 0.941 & 0.986 & 0.958 & 0.970 & 0.996 & 0.982 \\
  \multirow{1}*{MSNEA \citep{chen2022multi}} & 0.859 & 0.936 & 0.887 & 0.921 & 0.970 & 0.939 & 0.954 & 0.989 & 0.968 \\
  \multirow{1}*{MCLEA \citep{lin2022multi}} & 0.860 & 0.950 & 0.893 & 0.914 & 0.975 & 0.938 & 0.953 & 0.990 & 0.967 \\
  \multirow{1}*{MEAformer \citep{chen2023meaformer}} & 0.909 & 0.974 & 0.933 & 0.950 & 0.990 & 0.965 & 0.972 & 0.997 & 0.983 \\
  \cellcolor{cyan!10}\multirow{1}*{$\mathbf{\rm RA\text{-}MMEA\;(Ours)}$} & \cellcolor{cyan!10}$\mathbf{0.942}$ & \cellcolor{cyan!10}$\mathbf{0.992}$ & \cellcolor{cyan!10}$\mathbf{0.961}$ & \cellcolor{cyan!10}$\mathbf{0.971}$ & \cellcolor{cyan!10}$\mathbf{0.998}$ & \cellcolor{cyan!10}$\mathbf{0.982}$ & \cellcolor{cyan!10}$\mathbf{0.991}$ & \cellcolor{cyan!10}$\mathbf{0.999}$ & \cellcolor{cyan!10}$\mathbf{0.995}$ \\
  \cellcolor{cyan!5}\multirow{1}*{Gain over Prior Best (\%)} & \cellcolor{cyan!5}3.3 & \cellcolor{cyan!5}1.8 & \cellcolor{cyan!5}2.8 & \cellcolor{cyan!5}2.1 & \cellcolor{cyan!5}0.8 & \cellcolor{cyan!5}1.7 & \cellcolor{cyan!5}1.9 & \cellcolor{cyan!5}0.2 & \cellcolor{cyan!5}1.2 \\
  \toprule
  \multicolumn{10}{c}{$ \textit{\textbf{Iterative}} $}\\
  \toprule  
  \multirow{1}*{EVA \citep{liu2021visual}} & 0.948 & 0.992 & 0.964 & 0.977 & 0.997 & 0.985 & 0.990 & 1.000 & 0.995 \\
  \multirow{1}*{MSNEA \citep{chen2022multi}} & 0.871 & 0.943 & 0.898 & 0.927 & 0.976 & 0.945 & 0.959 & 0.993 & 0.979 \\
  \multirow{1}*{MCLEA \citep{lin2022multi}} & 0.942 & 0.994 & 0.963 & 0.974 & 0.999 & 0.984 & 0.986 & 1.000 & 0.992 \\
  \multirow{1}*{MEAformer \citep{chen2023meaformer}} & 0.964 & $0.998$ & 0.975 & 0.985 & 1.000 & 0.991 & 0.992 & 1.000 & 0.996 \\
  \cellcolor{cyan!10}\multirow{1}*{$\mathbf{\rm RA\text{-}MMEA\;(Ours)}$} & \cellcolor{cyan!10}$\mathbf{0.969}$ & \cellcolor{cyan!10}$\mathbf{0.999}$ & \cellcolor{cyan!10}$\mathbf{0.978}$ & \cellcolor{cyan!10}$\mathbf{0.989}$ & \cellcolor{cyan!10}$\mathbf{1.000}$ & \cellcolor{cyan!10}$\mathbf{0.993}$ & \cellcolor{cyan!10}$\mathbf{0.998}$ & \cellcolor{cyan!10}$\mathbf{1.000}$ & \cellcolor{cyan!10}$\mathbf{0.999}$ \\
  \cellcolor{cyan!5}\multirow{1}*{Gain over Prior Best (\%)} & \cellcolor{cyan!5}0.5 & \cellcolor{cyan!5}0.1 & \cellcolor{cyan!5}0.3 & \cellcolor{cyan!5}0.4 & \cellcolor{cyan!5}0.0 & \cellcolor{cyan!5}0.2 & \cellcolor{cyan!5}0.6 & \cellcolor{cyan!5}0.0 & \cellcolor{cyan!5}0.3 \\
  \bottomrule 
  \end{tabular}
  }
  \caption{\raggedright Unsupervised results on bilingual datasets are presented.
  }
  \label{tab:DBPUn}
\end{table*}

Tab.~\ref{tab:DBPUn} shows that our framework consistently outperforms all strong baselines on both Hits@1 and MRR, achieving improvements of 0.4\%–3.3\% and 0.2\%–2.8\%, respectively, across two settings on three bilingual datasets. Notably, the Hits@10 score reaches 1.0 on the $\rm DBP15K_{\emph{JA-EN}}$ and $\rm DBP15K_{\emph{FR-EN}}$.

\section{Appendix: Low-Resource Comparison}

To investigate the reliance of our framework on alignment labels during training, we evaluate RA-MMEA on FB15K-DB15K with fewer seed alignments. Following~\citet{chen2023meaformer} and ~\citet{li2025probing}, we set the seed alignment ratios to $R_{sa}$ = \{0.01, 0.03, 0.07, 0.11, 0.14, 0.18\}. For a more intuitive comparison of performance trends, we adopt the same backbone as their methods.

\begin{figure}[!t]
  \centering
  \includegraphics[width=0.49\linewidth]{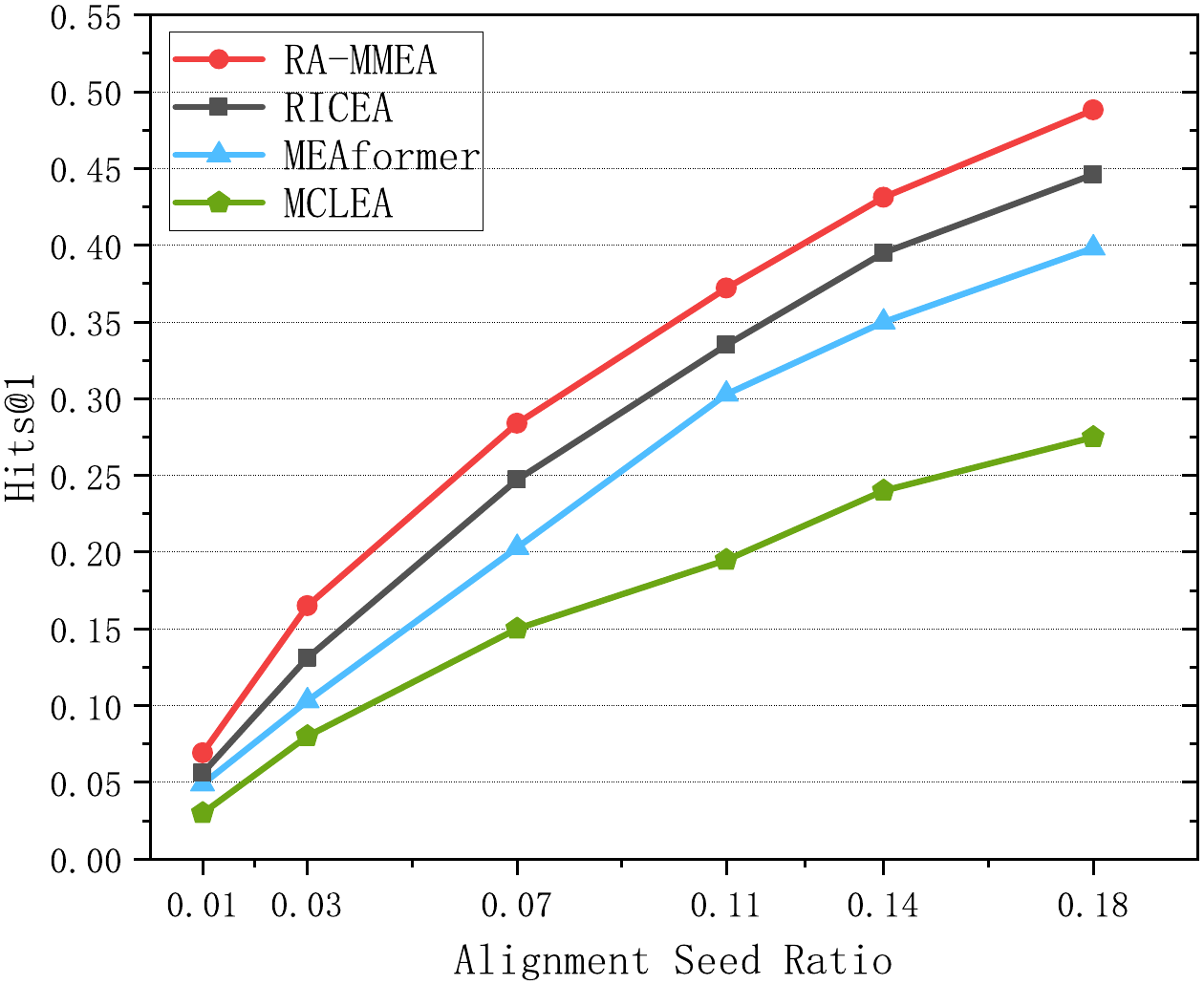}
  \includegraphics[width=0.49\linewidth]{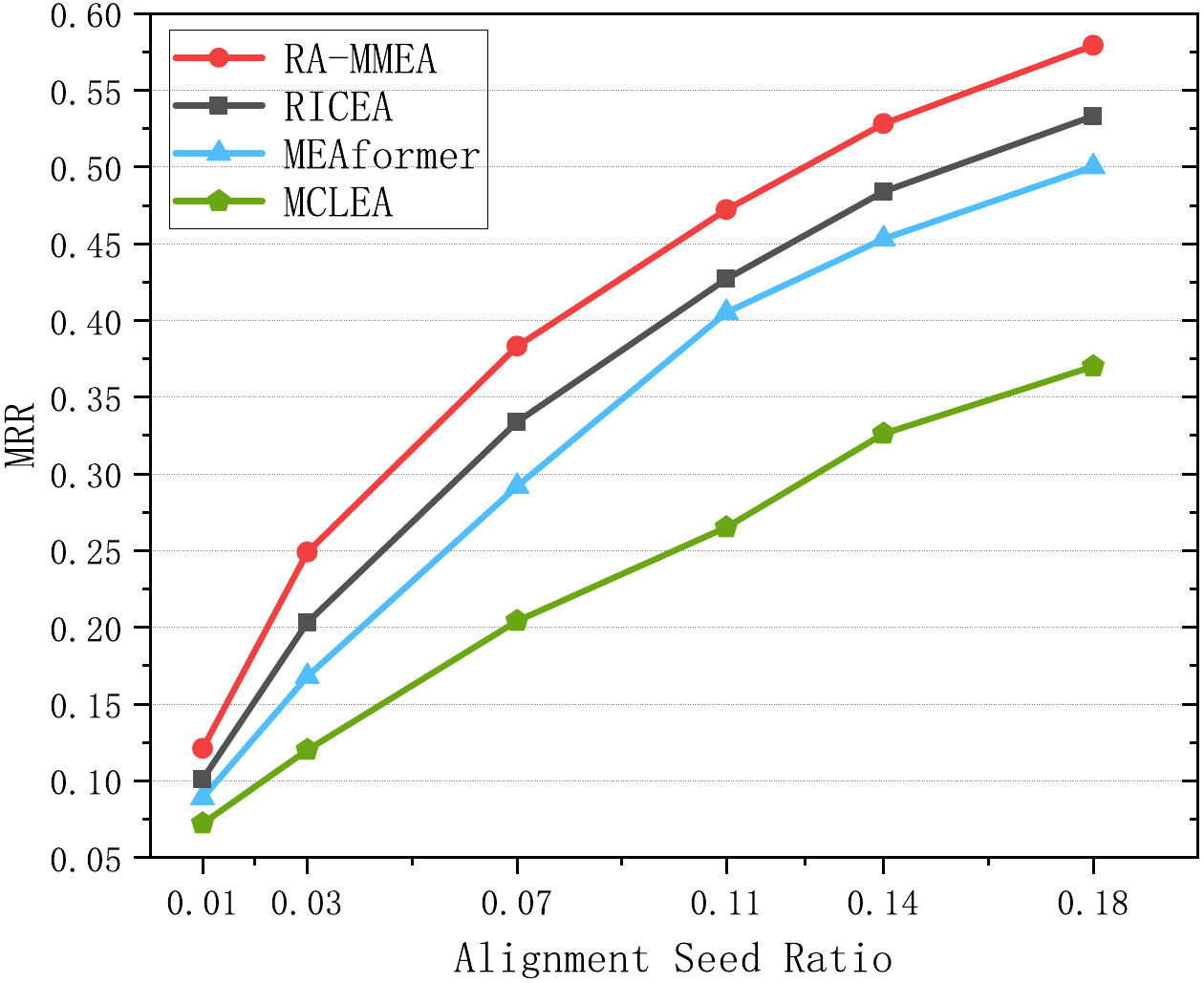}
  \caption{Performance of strong baselines with fewer seed alignments on FB15K-DB15K.}
  \label{fig:low}
\end{figure}

Fig.~\ref{fig:low} presents the comparison with other methods. RA-MMEA consistently outperforms all baselines across all seed alignment ratios, with increasingly pronounced advantages as more seed alignments become available. Notably, even when $R_{sa}=0.01$, RA-MMEA achieves the best Hits@1 and MRR among all compared methods. When $R_{sa}=0.18$, RA-MMEA even surpasses RICEA with $R_{sa}=0.20$. These results not only demonstrate the robustness of RA-MMEA to unreliable visual data but also highlight its potential in low-resource scenarios.

\section{Appendix: Applying DA-VRP to Textual Modalities}

To further evaluate DA-VRP’s ability to filter unreliable information beyond the visual modality, we additionally examine its performance on the relational and attribute modalities on FB15K-DB15K with 20\% seeds. In this section, we focus solely on identifying and removing unreliable textual modalities, without involving modality reconstruction. Please note that, we do not filter the structural modality, as it provides the essential backbone for modeling cross‑modal dependency. Removing structural information would compromise the stability of the alignment framework. 

\begin{table}[!t]
  \centering
  \renewcommand\arraystretch{1}
  \setlength{\tabcolsep}{12pt}
  \resizebox{\linewidth}{!}{
  \begin{tabular}{lccc}
    \hline
    \textbf{Models} & \textbf{Hits@1$\uparrow$} & \textbf{Hits@10$\uparrow$} & \textbf{MRR$\uparrow$} \\
    \hline
    RA-MMEA-R & 0.708 & 0.824 & 0.745 \\
    RA-MMEA-A & 0.732 & 0.845 & 0.811 \\
    \hline
    RA-MMEA-V & 0.798 & 0.886 & 0.855 \\
    \hline
  \end{tabular}
  }
  \captionsetup{justification=justified, singlelinecheck=false}
  \caption{Performance of applying our reliability prediction method to textual modalities.}
  \label{tab:gate4textual}
\end{table}

Tab.~\ref{tab:gate4textual} reports the results. RA-MMEA-V, RA-MMEA-R, and RA-MMEA-A respectively represent the variants applying reliability prediction on the visual image as proposed in this work, the relation modality, and the attribute modality. Please note that, for all these three variants, we do not perform modality reconstruction. As shown in Tab.~\ref{tab:gate4textual}, DA-VRP continues to provide clear benefits when applied to textual modalities, outperforming all other baselines in Tab.~\ref{tab:mono20seeds}. However, textual modalities often contain complementary relational cues, and aggressive filtering may inadvertently discard useful information. In future work, we will develop finer‑grained criteria to better separate complementary from redundant cross‑modal relationships, enabling more selective modality filtering. 

\section{Appendix: Visual Replacement Strategy} 

Although unreliable visual representations may still contain useful entity-related information, retaining them may also introduce additional noise. We therefore compare two alternative strategies that preserve the original unreliable visual representations: Concatenation and Weighted-Concatenation. Specifically, Concatenation combines the unreliable visual representation with the generated virtual visual representation, while Weighted-Concatenation reduces its contribution according to the estimated reliability.

\begin{table}[!t]
  \centering
  \renewcommand\arraystretch{1.00}
  \setlength{\tabcolsep}{7pt}
  \resizebox{\linewidth}{!}{
  \begin{tabular}{clccc}
    \hline
    &  & \textbf{Hits@1$\uparrow$} & \textbf{Hits@10$\uparrow$} & \textbf{MRR$\uparrow$} \\
    \hline
    \ding{182} & Concatenation & 0.831 & 0.941 & 0.873 \\
    \ding{183} & Weighted-Concatenation & 0.835 & 0.944 & 0.876 \\
    \rowcolor{cyan!10}
    \ding{184} & Ours & 0.849 & 0.948 & 0.879 \\
    \hline
  \end{tabular}
  }
  \captionsetup{justification=justified, singlelinecheck=false}
  \caption{Comparison of strategies for handling unreliable visual representations.}
  \label{tab:unaliable}
\end{table}

As shown in Tab.~\ref{tab:unaliable}, Weighted-Concatenation slightly outperforms Concatenation (\ding{183} \emph{v.s.} \ding{182}), indicating that suppressing unreliable visual information can alleviate its negative impact. Nevertheless, our strategy (\ding{184}) achieves the best performance across all metrics, improving Hits@1 by 1.4 and 1.8 percentage points over Weighted-Concatenation and Concatenation, respectively. These results suggest that retaining unreliable visual representations may still introduce misleading visual semantics into multi-modal fusion and consequently affect entity representations during the alignment process, especially when visual information is highly unreliable. In contrast, directly replacing them with virtual visual representations generated from the remaining reliable modalities provides more semantically consistent and informative visual representations for accurate entity alignment, validating the visual replacement strategy in RA-MMEA.

\section{Appendix: Model Complexity Analysis}

To further evaluate our framework, we compare the complexity of RA-MMEA with two representative lightweight baselines on FB15K-DB15K. As shown in Tab.~\ref{tab:Complexity}, the training time, learnable parameters, and GPU memory of our RA-MMEA are 24 minutes, 13M, and 16G, which are comparable to the complexities of MEAformer and RICEA. However, compared to these two methods, RA-MMEA achieves superior alignment performance on different metrics (see Tab.~\ref{tab:mono20seeds}, Appendix~\ref{sec:More Results} and Appendix~\ref{sec:Unsupervised Training}), showing an effective balance between accuracy and model complexity.

\begin{figure*}[!t]
  \centering
  \includegraphics[width=1.00\textwidth, keepaspectratio]{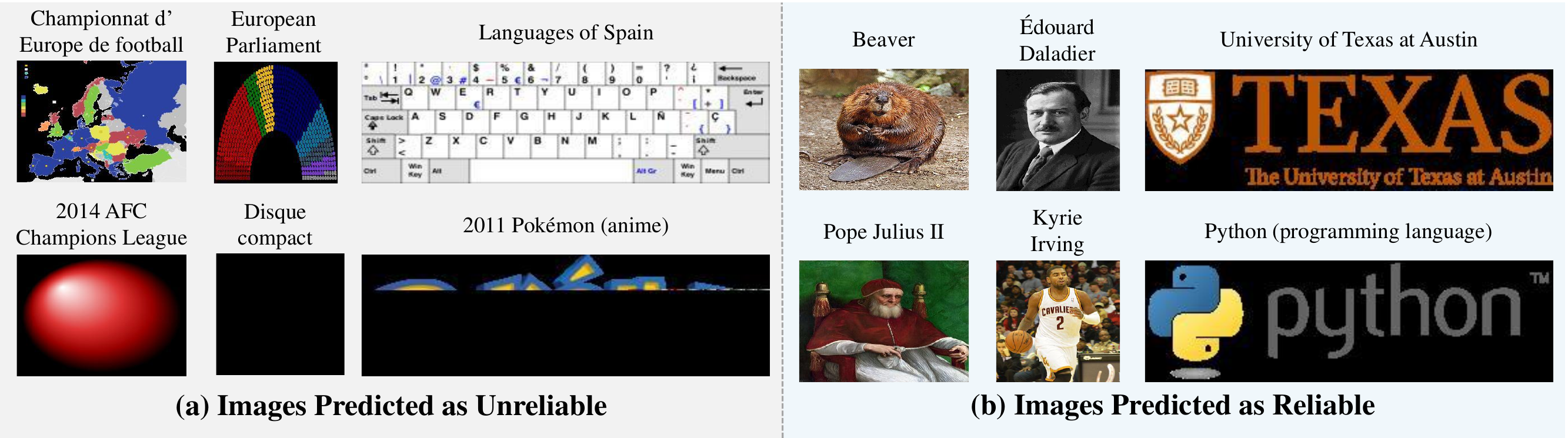}
  \caption{Case study of visual reliability on DBP15K, showing representative images predicted as unreliable (left) and reliable (right) by DA-VRP based on their cross-modal dependency with the corresponding entity semantics. The entity associated with each image is shown above the image.
  }
  \label{fig:case_study}
\end{figure*}

\begin{table}[!t]
  \centering
  \renewcommand\arraystretch{1.1}
  \setlength{\tabcolsep}{3pt}
  \resizebox{\linewidth}{!}{
  \begin{tabular}{lccc} 
  \hline
  \textbf{Models} & \textbf{\tabincell{c}{Train.\\ Time}} & \textbf{\tabincell{c}{Learnable \\ Paras.}} & \textbf{\tabincell{c}{GPU \\Mem.}} \\ 
  \hline
  MEAformer~\citep{chen2023meaformer}* & 22min & 10M  & 13.6G \\
  RICEA~\cite{li2025probing}* & 22min & 10M & 13.2G  \\
  RA-MMEA (Ours) & 24min & 13M & 16.0G  \\
  \hline 
  \end{tabular}
  }
  \captionsetup{justification=justified, singlelinecheck=false}
  \caption{Comparison of  complexity on FB15K-DB15K. * indicates our reproduced results.
  }
  \label{tab:Complexity}
\end{table}

\section{Appendix: Case Study}
\label{Appendix: Case Study}

To qualitatively examine whether DA-VRP can distinguish unreliable from reliable visual information, we present cases from DBP15K in Fig.~\ref{fig:case_study}. The left side shows images predicted as unreliable by DA-VRP. For \textit{European Parliament}, the image is an abstract seating diagram and provides limited visual semantics for representing the entity. The image associated with \textit{Compact disc} is nearly blank, while \textit{Championnat d'Europe de football} is represented by a general map of Europe rather than the football competition itself. Similar cases include \textit{2014 AFC Champions League}, \textit{2011 Pokémon (anime)}, and \textit{Languages of Spain}, whose images provide ambiguous, incomplete, or weakly related visual cues. These cases show that unreliable images may be irrelevant or capture only limited or indirect entity semantics.

In contrast, the reliable cases on the right exhibit entity--image semantic consistency. \textit{Édouard Daladier} is depicted by a portrait, \textit{Beaver} by an image of the corresponding animal, and \textit{University of Texas at Austin} by its recognizable institutional identity. Likewise, the images of \textit{Python (programming language)}, \textit{Pope Julius II}, and \textit{Kyrie Irving} contain distinctive visual information that directly corresponds to their entity semantics. Compared with the unreliable cases, these images provide more explicit and discriminative visual evidence for identifying the entities. The comparison shows that unreliable images may provide uninformative or weakly related semantics, potentially introducing misleading evidence, whereas reliable images exhibit stronger entity-image semantic consistency.

Notably, these examples cover diverse entity types, suggesting that the predicted reliability is not restricted to entity or visual content types. Despite their different visual characteristics, reliable images exhibit clearer semantic correspondence, while unreliable images provide more ambiguous or incomplete visual evidence. The qualitative results show that DA-VRP captures differences in visual reliability through cross-modal dependency. Together with the quantitative results in Sec.~\ref{Visual Noise Detection}, these cases further support cross-modal dependency as an effective task-level proxy for visual reliability.

\end{document}